\documentclass[11pt,a4paper]{article}
\usepackage[hyperref]{naaclhlt2018}
\usepackage{times}
\usepackage{latexsym}

\aclfinalcopy 
\def\aclpaperid{172} 

\usepackage{url}
\usepackage[utf8x]{inputenc}
\usepackage[pdftex]{graphicx}
\usepackage[all]{xy}
\usepackage{amsmath,amssymb,multirow,color}
\usepackage{algorithmic}
\usepackage[algoruled,linesnumbered,noend,noline]{algorithm2e}
\usepackage{multirow}

\usepackage{etoolbox}
\apptocmd{\thebibliography}{\raggedright}{}{}

\usepackage{dashrule}

\usepackage{rotating}
\usepackage{cleveref}

\title{Identifying Intensity of the Structure and Content in Tweets and the Discriminative Power of Attributes in Context with Referential Translation Machines}

\author{Ergun Bi\c{c}ici \\
            AI Enablement \\
            Huawei Türkiye R\&D Center \\
            Istanbul, Turkey \\
              \url{orcid.org/0000-0002-2293-2031} \\
              {\tt \url{ergun.bicici@huawei.com}} \\
              {\tt \url{bicici.github.io}}
}

\date{}

\begin{document}

\maketitle


\begin{abstract}
We use referential translation machines (RTMs) to identify the similarity between an attribute and two words in English by casting the task as machine translation performance prediction (MTPP) between the words and the attribute word and the distance between their similarities for Task 10 with stacked RTM models.
RTMs are also used to predict the intensity of the structure and content in tweets in English, Arabic, and Spanish in Task 1 
where MTPP is between the tweets and the set of words for the emotion selected from WordNet affect emotion lists.
Stacked RTM models obtain encouraging results in both.
\end{abstract}

\section{Introduction}

SemEval-2018~\cite{semeval2018} contained two prediction tasks that referential translation machine (RTM) models~\cite{Bicici:RTM:WMT2017,Bicici:RTM:SEMEVAL2017,Bicici:RTM_SEMEVAL} can be applied to enable new modeling capabilities and provide new results for comparison. The trace of the shared information with tweets or social media to the audiences' world can be approached by the intensity of the content and structure used in text with Task 1~\cite{semeval2018_Task1_affect_in_tweets}. The effectiveness of attributes to semantically separate two words from each other is the goal in Task 10~\cite{semeval2018_Task10_discriminative_attributes}, which can be used to improve natural language understanding systems. Table~\ref{DataSize} lists the number of instances in the provided datasets. We use RTMs to model both tasks.

\begin{figure}[t]
\centering
\includegraphics[width=1.0\linewidth]{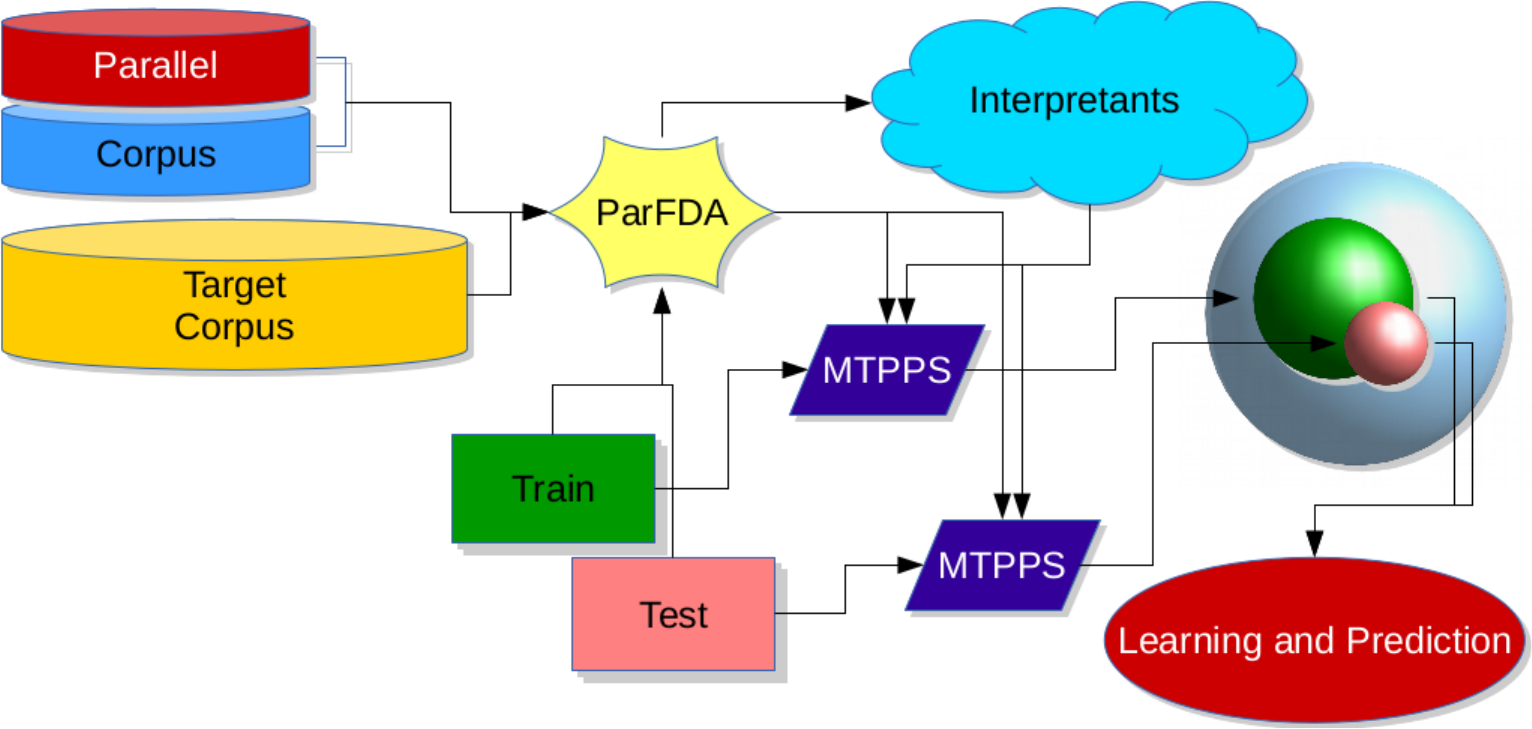}
\caption{RTM depiction: \texttt{parfda} selects interpretants close to the data using corpora; 
two MTPPS use interpretants, training data, and test data to generate features in the same space; 
learning and prediction use these features as input. Spheres are for feature spaces.}
\label{RTMDiagram}
\end{figure}

RTMs use \texttt{parfda}~\cite{Bicici:ParFDA:WMT2016} to select both parallel and monolingual interpretants, data close to the task instances selected specifically for the task, to derive features measuring the closeness of the test sentences to the training data, the difficulty of translating them, and to identify translation acts between any two data sets using machine translation performance prediction system (MTPPS)~\cite{Bicici:MTPP:MTJ2013,Bicici:MTPPS:SNCS2022} to build prediction models. Interpretants provide context and text for feature derivation to link translation source and target and training and test sets.
RTMs are applicable in different domains and tasks and in both monolingual and bilingual settings. 
Figure~\ref{RTMDiagram} explains RTMs' model building process. 

\section{Predicting the Discriminative Power of Attributes}

Capturing discriminative attributes task (Task 10)~\cite{semeval2018_Task10_discriminative_attributes}
is looking at whether an attribute (e.g. red) can be used to discriminate between two other words (e.g. apple and banana) to complement semantic similarity efforts. The answer to whether the attribute can be used to discriminate the two words can be useful for semantic similarity with contextual dependency. 
The task is posed as a binary classification task: $f(w_1, w_2, a) \rightarrow 0 \; ? \; 1$.
The target value to predict is either $0$ or $1$ and shows whether the attribute can be used for discrimination for the given context. Evaluation metric is $F_1$.
RTMs are used by casting the task as MTPP between the words and the attribute and building predictors that use the distance between the predictions. We assume that the discriminative power increase when the attribute is similar to words with significant difference.

\begin{table}[t]
{
\begin{center}
\begin{tabular}{@{\hspace{0.0cm}}c@{\hspace{0.1cm}}c@{\hspace{0.1cm}}|c@{\hspace{0.1cm}}c@{\hspace{0.1cm}}c@{\hspace{0.1cm}}c@{\hspace{0.1cm}}c@{\hspace{0.0cm}}}
& & & English & Arabic & Spanish \\
train & Task 1 & emotion & 7102 & 3376 & 4544 \\
dev & Task 1 & emotion & 1464 & 661 & 793 \\
test & Task 1 & emotion & 71816 & 1563 & 2616 \\
train & Task 1 & valence & 1181 & 932 & 1566 \\
dev & Task 1 & valence & 449 & 138 & 229 \\
test & Task 1 & valence & 17874 & 730 & 648 \\
\hline
train & Task 10 & & 17521 \\
dev & Task 10 & & 2722 \\
test & Task 10 & & 2340 \\
\hline
\end{tabular}
\begin{tabular}{@{\hspace{0.0cm}}c@{\hspace{0.1cm}}c@{\hspace{0.1cm}}|c@{\hspace{0.1cm}}c@{\hspace{0.1cm}}c@{\hspace{0.1cm}}c@{\hspace{0.1cm}}c@{\hspace{0.0cm}}}
Task 1 & emotion & English & Arabic & Spanish \\
\hline
train & anger & 1701 & 877 & 1166 \\
train & fear & 2252 & 882 & 1166 \\
train & joy & 1616 & 728 & 1058 \\
train & sadness & 1533 & 889 & 1154 \\
dev & anger & 388 & 150 & 193 \\
dev & fear & 389 & 146 & 202 \\
dev & joy & 290 & 224 & 202 \\
dev & sadness & 397 & 141 & 196 \\
test & anger & 17939 & 373 & 627 \\
test & fear & 17923 & 372 & 618 \\
test & joy & 18042 & 448 & 730 \\
test & sadness & 17912 & 370 & 641 \\
\hline
\end{tabular}
\end{center}
}
\caption{Number of instances in the provided data. About $95\%$ of the Task 1 emotion English test sets are not evaluated for the competition rankings and 1002, 986, 1105, and 975 instances were actually scored.}
\label{DataSize}
\end{table}

\begin{figure*}[t]
\centering
\includegraphics[width=1.0\linewidth]{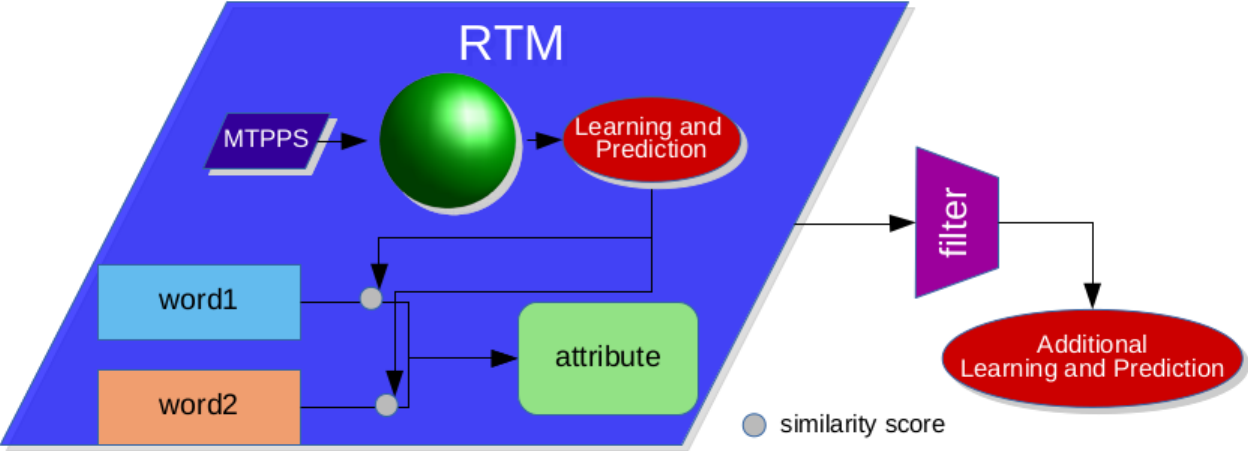}
\caption{RTM with stacked combined prediction use a combined model to obtain feature representations and predictions for both $w_1 \rightarrow a$ and $w_2 \rightarrow a$, which are processed before additional learning and prediction.}
\label{RTM_combined_predictor}
\end{figure*}

The stacked RTM model with combined prediction step (\Cref{RTM_combined_predictor}) use the same model to predict the MTPP similarity of $w_1 \rightarrow a$ and of $w_2 \rightarrow a$ where the data is collected such that the first row for each attribute is for $w_1 \rightarrow a$ and the second is for $w_2 \rightarrow a$. Stacking is used to build higher level models using predictions from base prediction models where they can also use the probability associated with the predictions~\cite{Stacking1999}. The combined model in \Cref{RTM_combined_predictor} is adding the predictions as additional feature.
We obtain an RTM representation vector for each of the instances by using the derived features and the combination of the prediction scores along with 5 additional features:
\begin{align}
\hat{y}_1 & & \hat{y}_2 & & |\hat{y}_1 - \hat{y}_2| & & (\hat{y}_1 + \hat{y}_2) / 2 & & \sqrt{\hat{y}_1*\hat{y}_2}
\end{align}
After this filtering step, we run another learning and prediction on the concatenation of the features from both rows and the additional features.

\begin{figure*}[t]
\centering
\includegraphics[width=1.0\linewidth]{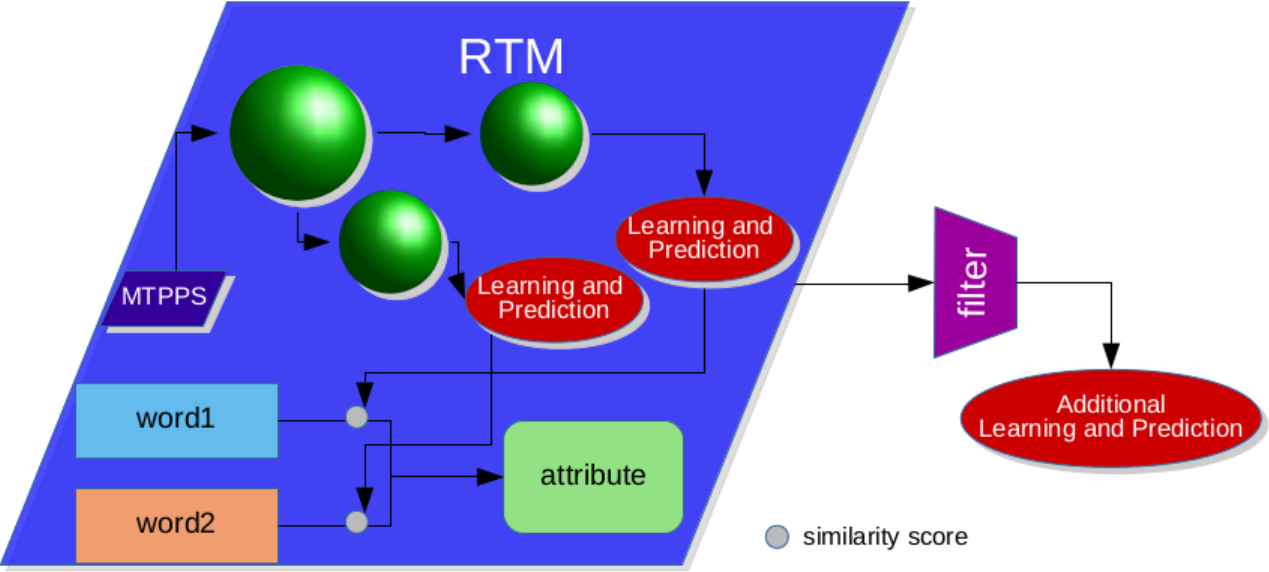}
\caption{RTM with stacked separate predictions use two different learning steps to obtain feature representations and predictions for either $w_1 \rightarrow a$ or $w_2 \rightarrow a$, which are processed before additional learning and prediction.}
\label{RTM_separate_predictors}
\end{figure*}

The stacked RTM model with separate prediction steps use separated feature sets with an initial prediction with each (\Cref{RTM_separate_predictors}). 
Using separate predictors for each word need not achieve better performance than using a single predictor since the rows of words we compare need not belong to different groups due to their ordering in the dataset and the words may be only randomly positioned. However, both predictions can be useful as coming from two weaker predictors since half of the training instances were used to train each. The benefit of using separate feature sets is further specialization of $w_1 \rightarrow a$ and $w_2 \rightarrow a$ models with feature selection and partial-least squares from \Cref{RTMResults}.
As we see in \Cref{TestResultsNew}, separated feature sets with two separate learning and prediction steps improve the performance. 



The architectures of \Cref{RTM_combined_predictor} and \Cref{RTM_separate_predictors} are general enough to be useful for Task 1 as well and when we want to make use of the differences between the predictions for the same instances. 

\section{Predicting the Intensity of the Structure and Content in Tweets}

Affect in tweets~\footnote{\url{www.twitter.com}} task (Task 1)~\cite{semeval2018_Task1_affect_in_tweets} is about predicting the intensity of the emotion expressed for tweets within sadness, joy, fear, or anger emotion categorizations or the valance (sentiment). The emotion within tweets is about how the tweeter wrote and valence or sentiment is about what the tweeter wrote. 
Intensity scores are obtained with best-worst scaling (bws)~\cite{LREC18-TweetEmo}, which counts only the number of times a tweet is labeled as best or worst among 4 tweets where each is annotated by multiple workers. The scores are obtained with the percentage of counts scaled to $[-1, 1]$, which is later scaled to $[0, 1]$ for the task. bws can decrease the annotation effort to obtain the set of binary comparisons used to obtain reliably agreed labels.\footnote{bws is similar to inter-annotator agreement (IAA) $\tau$, which is in $[-1, 1]$ and calculated as IAA $\tau = (C - D) / (C+D)$, where C is the number of concordant pairs and D is the number of discordant pairs. IAA $\tau$ is the same measure as ``Kendall's $\tau$ with ties penalized'' used in~\cite{WMT2013} (Task1.2), where it was used for measuring the correlation between quality estimation systems and human rankings.
Oracle METEOR evaluation achieved $\tau=0.23$ for ranking in 2013's quality estimation task (Task 1.2, see \cite{WMT2013}). 
We tried a randomized IAA (RIAA) $\tau$~\cite{QTLeap} where the calculations are similar to IAA $\tau$, but all ties were converted into \emph{better} and \emph{worse} ranking randomly to obtain more robust results by distributing the ambiguous counts. This also helps normalization of the counts such that their sum are same for different models. A corresponding randomized bws would randomly assign the remaining pairwise comparison counts, where for 4 tweets only a single comparison would be left ambiguous, and the count can be distributed evenly.}

We model the task as MTPP of the tweets to the emotions to answer questions like ``to what degree is this tweet showing the emotion of''. Since a single emotion word need not provide enough context for semantic discrimination, we use sets of words for each emotion that express the same meaning using a subset of the WordNet affect emotion lists~\cite{Strapparava2004}. 
The lexicon used for English is in \Cref{wordnetlexicon}. We obtained their translations to Arabic and Spanish using web translation sites~\footnote{e.g. \url{translate.google.com} or \url{www.bing.com/translator}} to obtain the corresponding lexicon.
We use the whole set of words corresponding to the emotion instead of the emotion word to translate to.
For valence intensity tasks, we used both of the sets of words from emotions joy and sadness as a single sentence to translate to and we also used them as separate rows using \Cref{RTM_combined_predictor} and \Cref{RTM_separate_predictors} where the tweet's MTPP is predicted to the sets of words of either joy or sadness (\Cref{Task1_RTMseparated}).
We participate in the regression tasks of Task 1 to predict the emotion and valence intensity in Arabic, English, and Spanish tweets. The official evaluation metric is Pearson's correlation.~\footnote{The program for evaluation is at \url{https://github.com/felipebravom/SemEval_2018_Task_1_Eval}}

\begin{table*}[t]
{
\begin{center}
\begin{tabular}{@{\hspace{0.0cm}}llll|llllll@{\hspace{0.0cm}}}
\multicolumn{3}{c}{Task} & & $r$ & MAE & RAE & MAER & MRAER \\
\hline
English & Task 1 & emotion & anger & 0.104 & 0.1744 & 1.121 & 0.3849 & 1.106 \\
English & Task 1 & emotion & fear & 0.063 & 0.169 & 1.156 & 0.3869 & 1.155 \\
English & Task 1 & emotion & joy & 0.266 & 0.1859 & 1.193 & 0.3849 & 1.293 \\
English & Task 1 & emotion & sadness & 0.233 & 0.1673 & 1.114 & 0.3543 & 1.165 \\
English & Task 1 & emotion & ALL & 0.168 & 0.1745 & 1.147 & 0.3781 & 1.182 \\
Arabic & Task 1 & emotion & anger & 0.196 & 0.1407 & 0.988 & 0.3112 & 0.886 \\
Arabic & Task 1 & emotion & fear & 0.125 & 0.1445 & 1.01 & 0.3152 & 0.904 \\
Arabic & Task 1 & emotion & joy & 0.316 & 0.1431 & 0.949 & 0.4076 & 0.826 \\
Arabic & Task 1 & emotion & sadness & 0.238 & 0.1452 & 0.99 & 0.3238 & 0.89 \\
Arabic & Task 1 & emotion & ALL & 0.229 & 0.1434 & 0.981 & 0.3428 & 0.871 \\
Spanish & Task 1 & emotion & anger & 0.211 & 0.171 & 1.007 & 0.3796 & 0.957 \\
Spanish & Task 1 & emotion & fear & 0.35 & 0.1612 & 0.923 & 0.4543 & 0.837 \\
Spanish & Task 1 & emotion & joy & 0.389 & 0.1596 & 0.9 & 0.471 & 0.838 \\
Spanish & Task 1 & emotion & sadness & 0.415 & 0.1491 & 0.896 & 0.3968 & 0.833 \\
Spanish & Task 1 & emotion & ALL & 0.34 & 0.1602 & 0.928 & 0.4273 & 0.858 \\
ALL & Task 1 & emotion & ALL & 0.216 & 0.1641 & 1.041 & 0.3878 & 1.021 \\
\hline
English & Task 1 & valence & & 0.161 & 0.2642 & 1.458 & 0.5155 & 1.674 \\
Arabic & Task 1 & valence & & 0.264 & 0.1921 & 0.958 & 0.5595 & 0.919 \\
Spanish & Task 1 & valence & & 0.272 & 0.1665 & 0.962 & 0.5302 & 0.85 \\
ALL & Task 1 & valence & & 0.117 & 0.2141 & 1.154 & 0.5333 & 1.185 \\
\hline
English & Task 10 & \multicolumn{2}{c|}{\multirow{3}{*}{combined}} & -0.045 & 0.5034 & 1.018 & 1.0954 & 0.994 \\
& & & & $F_1$ & \\ 
English & Task 10 & & & 0.47 & \\ 
\hline
\end{tabular}
\end{center}
}
\caption{Competition results on the test set.}
\label{TestResults}
\end{table*}


\begin{table*}[t]
{
\small
\begin{center}
\begin{tabular}{@{\hspace{0.0cm}}c@{\hspace{0.1cm}}|c@{\hspace{0.1cm}}c@{\hspace
{0.1cm}}c@{\hspace{0.1cm}}c@{\hspace{0.1cm}}c@{\hspace{0.1cm}}c@{\hspace{0.1cm}}
c@{\hspace{0.1cm}}c@{\hspace{0.0cm}}}
& T1 EI-en & T1 EI-ar & T1 EI-es & T1 V-en & T1 V-ar & T1 V-es & T10 \\
\hline
ranks & 44 & 13 & 10 & 35 & 13 & 10 & 9 \\
out of & 50 & 15 & 17 & 39 & 15 & 15 & 9 \\
\hline
\end{tabular}
\end{center}
}
\caption{RTM ranks at SemEval-2018.}
\label{TestsetRanks}
\end{table*}

\begin{table*}[t]
{
\begin{center}
\begin{tabular}{@{\hspace{0.0cm}}llll|lllll@{\hspace{0.0cm}}}
\multicolumn{3}{c}{Task} & & $r$ & MAE & RAE & MAER & MRAER \\ 
\hline
English & Task 1 & emotion & anger & 0.245 & 0.1689 & 1.086 & 0.3619 & 1.074 \\ 
English & Task 1 & emotion & fear & 0.05 & 0.1526 & 1.044 & 0.413 & 0.952 \\ 
English & Task 1 & emotion & joy & 0.028 & 0.1641 & 1.053 & 0.438 & 0.963 \\ 
English & Task 1 & emotion & sadness & 0.004 & 0.1566 & 1.043 & 0.4064 & 0.944 \\ 
English & Task 1 & emotion & ALL & 0.2245 & 0.1734 & 1.14 & 0.367 & 1.1693 \\ 
Arabic & Task 1 & emotion & anger & 0.209 & 0.1413 & 0.992 & 0.3093 & 0.878 \\ 
Arabic & Task 1 & emotion & fear & 0.173 & 0.1444 & 1.01 & 0.3112 & 0.905 \\ 
Arabic & Task 1 & emotion & joy & 0.377 & 0.1417 & 0.94 & 0.4126 & 0.805 \\ 
Arabic & Task 1 & emotion & sadness & 0.269 & 0.1442 & 0.983 & 0.3291 & 0.872 \\ 
Arabic & Task 1 & emotion & ALL & 0.2543 & 0.1428 & 0.9771 & 0.344 & 0.8545 \\ 
Spanish & Task 1 & emotion & anger & 0.183 & 0.1706 & 1.005 & 0.3807 & 0.927 \\ 
Spanish & Task 1 & emotion & fear & 0.398 & 0.1607 & 0.92 & 0.4548 & 0.846 \\ 
Spanish & Task 1 & emotion & joy & 0.298 & 0.1676 & 0.945 & 0.4856 & 0.843 \\ 
Spanish & Task 1 & emotion & sadness & 0.405 & 0.1513 & 0.909 & 0.3943 & 0.838 \\ 
Spanish & Task 1 & emotion & ALL & 0.324 & 0.1627 & 0.9426 & 0.4311 & 0.8568 \\ 
\hline
English & Task 1 & valence & & 0.1326 & 0.1884 & 1.0399 & 0.525 & 0.9791 \\ 
Arabic & Task 1 & valence & & 0.2981 & 0.1879 & 0.9366 & 0.5637 & 0.8482 \\ 
Spanish & Task 1 & valence & & 0.2152 & 0.1684 & 0.973 & 0.5399 & 0.8317 \\ 
\hline
\end{tabular}
\end{center}
}
\caption{RTM results on the test set after the competition.}
\label{TestResultsNew}
\end{table*}

\begin{table*}[t]
{
\begin{center}
\begin{tabular}{@{\hspace{0.0cm}}llll|lllll@{\hspace{0.0cm}}}
\multicolumn{3}{c}{Task} & & $r$ & MAE & RAE & MAER & MRAER \\ 
\hline
English & Task 1 & emotion & anger & 0.245 & 0.1689 & 1.086 & 0.3619 & 1.074 \\ 
English & Task 1 & emotion & fear & 0.05 & 0.1526 & 1.044 & 0.413 & 0.952 \\ 
English & Task 1 & emotion & joy & 0.028 & 0.1641 & 1.053 & 0.438 & 0.963 \\ 
English & Task 1 & emotion & sadness & 0.004 & 0.1566 & 1.043 & 0.4064 & 0.944 \\ 
English & Task 1 & emotion & ALL & 0.2245 & 0.1734 & 1.14 & 0.367 & 1.1693 \\ 
Arabic & Task 1 & emotion & anger & 0.209 & 0.1413 & 0.992 & 0.3093 & 0.878 \\ 
Arabic & Task 1 & emotion & fear & 0.177 & 0.1443 & 1.009 & 0.311 & 0.905 \\ 
Arabic & Task 1 & emotion & joy & 0.377 & 0.1416 & 0.939 & 0.4123 & 0.805 \\ 
Arabic & Task 1 & emotion & sadness & 0.271 & 0.144 & 0.982 & 0.329 & 0.871 \\ 
Arabic & Task 1 & emotion & ALL & 0.2562 & 0.1428 & 0.9765 & 0.3439 & 0.8541 \\ 
Spanish & Task 1 & emotion & anger & 0.183 & 0.1706 & 1.004 & 0.3807 & 0.926 \\ 
Spanish & Task 1 & emotion & fear & 0.398 & 0.1607 & 0.921 & 0.4551 & 0.846 \\ 
Spanish & Task 1 & emotion & joy & 0.297 & 0.1677 & 0.946 & 0.4857 & 0.842 \\ 
Spanish & Task 1 & emotion & sadness & 0.405 & 0.1514 & 0.91 & 0.3944 & 0.837 \\ 
Spanish & Task 1 & emotion & ALL & 0.3235 & 0.1627 & 0.9427 & 0.4313 & 0.8565 \\ 
\hline
English & Task 1 & valence & & 0.1332 & 0.1888 & 1.042 & 0.5254 & 0.9842 \\ 
Arabic & Task 1 & valence & & 0.3023 & 0.1874 & 0.9344 & 0.5628 & 0.8471 \\ 
Spanish & Task 1 & valence & & 0.2149 & 0.1684 & 0.973 & 0.5399 & 0.8321 \\ 
\hline
\end{tabular}
\end{center}
}
\caption{RTM results on the test set after the competition with symbolic grounding and scaling of predictions.}
\label{TestResultsNew2}
\end{table*}

\begin{table*}[t]
{
\small
\begin{center}
\begin{tabular}{@{\hspace{0.0cm}}llll|lllllll@{\hspace{0.0cm}}}
\multicolumn{3}{c}{Task} & & $r$ & MAE & RAE & MAER & MRAER & rMAER & rMRAER \\ 
\hline
without & MIX \\
English & Task 1 & emotion & ALL & 0.2106 &  0.1755  & 1.1536 &  0.3687  & 1.1936  & 0.0795 &  0.4261 \\
Arabic & Task 1 & emotion & ALL & 0.256 & 0.1428 & 0.9771 &  0.3442 &  0.852 & 0.0515 & 0.2447 \\
Spanish & Task 1 & emotion & ALL & 0.3125 & 0.163 & 0.9445 & 0.432 & 0.8605 & 0.0564 & 0.2315 \\
English & Task 1 & valence & & 0.1203 & 0.1841 & 1.0164 & 0.5569 & 0.9296 & 0.0615 & 0.2367 \\
Arabic & Task 1 & valence & & 0.225 & 0.1927 & 0.9608 & 0.5702 & 0.8984 & 0.0615 & 0.1952 \\
Spanish & Task 1 & valence & & 0.2216 & 0.1683 & 0.9724 & 0.5416 & 0.8357 & 0.0533 & 0.187 \\
\hline
with & MIX \\
English & Task 1 & emotion & ALL & 0.2351 & 0.1693 & 1.1129 & 0.3648 & 1.1204  & 0.0722 & 0.3857 \\
Arabic & Task 1 & emotion & ALL & 0.2543 & 0.1428 & 0.9771 & 0.344 & 0.8545 & 0.051 & 0.2482 \\
Spanish & Task 1 & emotion & ALL & 0.324 & 0.1627 & 0.9426 & 0.4311 & 0.8568 & 0.0548 & 0.2298 \\
English & Task 1 & valence & & 0.1326 & 0.1884 & 1.0399 & 0.525 & 0.9791 & 0.0674 & 0.2808 \\
Arabic & Task 1 & valence & & 0.2981 & 0.1879 & 0.9366 & 0.5637 & 0.8482 & 0.0604 & 0.198 \\
Spanish & Task 1 & valence & & 0.2169 & 0.1683 & 0.9726 & 0.5396 & 0.8313 & 0.052 & 0.19 \\
\hline
with & MIX & symbolic & grounding \\
English & Task 1 & emotion & ALL & 0.2105 & 0.1738 & 1.1423 & 0.3675 & 1.1705 & 0.0768 & 0.4116 \\
Arabic & Task 1 & emotion & ALL & 0.2546 & 0.1429 & 0.9772 & 0.3441 & 0.8542 & 0.051 & 0.248 \\
Spanish & Task 1 & emotion & ALL & 0.324 & 0.1627 & 0.9427 & 0.4313 & 0.8563 & 0.0547 & 0.2293 \\
English & Task 1 & valence & & 0.129 & 0.1893 & 1.045 & 0.5274 & 0.9891 & 0.0688 & 0.2826 \\
Arabic & Task 1 & valence & & 0.3023 & 0.1875 & 0.9345 & 0.5632 & 0.847 & 0.0619 & 0.2002 \\
Spanish & Task 1 & valence & & 0.2171 & 0.1683 & 0.9726 & 0.5397 & 0.831 & 0.0521 & 0.1901 \\
\hline
with & new MIX & symbolic & grounding \\
English & Task 1 & emotion & ALL & 0.2355 & 0.1692 & 1.112 & 0.3648 & 1.1187 & 0.0719 & 0.3848 \\
Arabic & Task 1 & emotion & ALL & 0.255 & 0.1428 & 0.977 & 0.3441 & 0.8528 & 0.0526 & 0.247 \\
Spanish & Task 1 & emotion & ALL & 0.3236 & 0.1627 & 0.9427 & 0.4311 & 0.8567 & 0.0543 & 0.2296 \\
English & Task 1 & valence & & 0.1324 & 0.1881 & 1.0381 & 0.5251 & 0.9751 & 0.0665 & 0.2783 \\
Arabic & Task 1 & valence & & 0.2983 & 0.1879 & 0.9366 & 0.5637 & 0.848 & 0.0602 & 0.1978 \\
Spanish & Task 1 & valence & & 0.2175 & 0.1683 & 0.9725 & 0.5396 & 0.8308 & 0.0519 & 0.1899 \\
\hline
with & fold MIX & symbolic & grounding \\
English & Task 1 & emotion & ALL & 0.2355 & 0.1692 & 1.1119 & 0.3647 & 1.1185 & 0.0717 & 0.3844 \\
Arabic & Task 1 & emotion & ALL & 0.2555 & 0.1428 & 0.9771 & 0.344 & 0.8533 & 0.0526 & 0.249 \\
Spanish & Task 1 & emotion & ALL & 0.324 & 0.1627 & 0.9426 & 0.4311 & 0.8568 & 0.055 & 0.2299 \\
English & Task 1 & valence & & 0.1321 & 0.1881 & 1.0381 & 0.5249 & 0.9752 & 0.0676 & 0.279 \\
Arabic & Task 1 & valence & & 0.2997 & 0.1877 & 0.9358 & 0.5638 & 0.8475 & 0.0597 & 0.1972 \\
Spanish & Task 1 & valence & & 0.2169 & 0.1683 & 0.9724 & 0.5397 & 0.831 & 0.0521 & 0.1898 \\
\end{tabular}
\end{center}
}
\caption{RTM results on the test set without symbolic grounding and scaling of predictions.}
\label{TestResultsNew3}
\end{table*}

\section{RTM Models and Results}
\label{RTMResults}

The RTM models used predictions from machine learning models including 
ridge regression (RR), k-nearest neighors (KNN), support vector regression (SVR), 
AdaBoost~\cite{AdaBoost}, and extremely randomized trees (TREE)~\cite{ExtrementRandomizedTrees} 
in combination with feature selection (FS)~\cite{GuyonWBV02} and partial least squares (PLS)~\cite{PLS1984}. 
We use averaging of scores from different models for robustness~\cite{Bicici:RTM:SEMEVAL2017}. 
Model implementations use \texttt{scikit-learn}.~\footnote{\url{http://scikit-learn.org/}}
We optimize $\lambda$ for RR, k for KNN, $\gamma$, C, and $\epsilon$ for SVR, minimum number of samples for leaf nodes and for splitting an internal node for TREE, the number of features for FS, and the number of dimensions for PLS. 
We use 500 estimators in the TREE model and also for AdaBoost along with exponential loss.
We use grid search for SVR. 
We evaluate with Pearson's correlation ($r$), mean absolute error 
(MAE), relative absolute error (RAE), MAER (mean absolute error relative), and 
MRAER (mean relative absolute error relative)~\cite{Bicici:RTM_SEMEVAL}.
We use $7$-fold cross-validation performance on the training set to rank models.

\begin{table}
\begin{tabular}{l}
$1$\&$2$-gram wrec \\
$1$\&$2$\&$3$-gram wGM \\
$1$\&$2$-gram wGM \\
$1$\&$2$-gram w$F_1$ \\
$1$-gram wGM
\end{tabular}
\caption{Top $5$ features selected for Task 10}
\label{top5features_Task10}
\end{table}

\begin{table}
\begin{tabular}{l}
translation logprobability bpw \\
word alignment (1 - WER) \\
word alignment $F_1$ score \\
$3$gram w$F_1$ \\
sentence number of characters \\
\end{tabular}
\caption{Top $5$ features selected for Task 1 Spanish}
\label{top5features_Task1}
\end{table}



Interpretants are selected from the corpora distributed by the translation task of WMT17~\cite{WMT2017} and they consist of monolingual sentences used to build the LM and parallel sentence pair instances used by MTPPS to derive the features. We built RTM models using:
\begin{itemize}
\item 300 thousand sentences for training data
\item 5 million sentences for LM
\end{itemize}
RTMs generate features for the training and the test set to map both to the same space where the total 
number of features in Task 1 becomes 492 and Task 10 becomes 117. 
The difference is due to the smaller context the attribute word provides and most of the sentence-level features become not useful including the sentence structure parsing features or word alignment features.

\begin{table*}[t]
{
\begin{center}
\begin{tabular}{@{\hspace{0.0cm}}l|c|lllll@{\hspace{0.0cm}}}
Setting & $F_1$ & $r$ & MAE & RAE & MAER & MRAER \\
\hline
combined & 0.5169 & 0.035 & 0.4735 & 0.958 & 1.1397 & 0.95 \\
\hline
separate & 0.5006 & 0.014 & 0.4761 & 0.963 & 1.0317 & 0.939 \\
\hline
\end{tabular}
\end{center}
}
\caption{Task 10 test set results with translation similarity distance modeling.}
\label{Task10_translation_similarity_distance}
\end{table*}

RTM results in the Task 1~\footnote{\url{https://competitions.codalab.org/competitions/17751}} and Task 10~\footnote{\url{https://competitions.codalab.org/competitions/17326}} competitions are in \Cref{TestResults} and ranks are in \Cref{TestsetRanks}~\cite{semeval2018_Task1_affect_in_tweets}. 
8 of the results obtain MRAER larger than 1 suggesting more work towards these tasks or subtasks. weight \# models combine top \# models' predictions~\cite{Bicici:RTM:SEMEVAL2017}.
The predictions for Task 10 were transformed to binary classes by thresholding with $0.5$ and obtains $0.47$ $F_1$. 

\begin{table*}[t]
{
\begin{center}
\begin{tabular}{@{\hspace{0.0cm}}lll|lllll@{\hspace{0.0cm}}}
\multicolumn{3}{c|}{Task 1} & $r$ & MAE & RAE & MAER & MRAER \\
\hline
English & \multicolumn{2}{l|}{tweet $\rightarrow$ $V_{joy} \cup V_{sadness}$} & 0.1271 & 0.1973 & 1.0892 & 0.5248 & 1.0739 \\ 
& \multicolumn{2}{l|}{tweet $\rightarrow$ $V_{joy}$} & 0.2121 & 0.1922 & 1.061 & 0.5334 & 1.0907 \\
& \multicolumn{2}{l|}{tweet $\rightarrow$ $V_{sadness}$} & 0.2869 & 0.2027 & 1.1188 & 0.4552 & 1.1848 \\
& \multicolumn{2}{l|}{combined} & 0.2536 & 0.1891 & 0.943 & 0.56 & 0.8583 \\
& \multicolumn{2}{l|}{separate} & 0.2779 & 0.2477 & 1.3671 & 0.4883 & 1.5812 \\
& \multicolumn{2}{l|}{separate+combined} & 0.269 & 0.1973 & 1.0892 & 0.5298 & 1.1598 \\
& \multicolumn{2}{l|}{separate+combined} & 0.3555 & 0.176 & 0.9714 & 0.4542  & 0.9547 (MRAER) \\
\hline
Arabic & \multicolumn{2}{l|}{tweet $\rightarrow$ $V_{joy} \cup V_{sadness}$} & 0.2398 & 0.1921 & 0.9576 & 0.5486 & 0.8903 \\ 
& \multicolumn{2}{l|}{tweet $\rightarrow$ $V_{joy}$} & 0.2127 & 0.1929 & 0.9616 & 0.565 & 0.8897 \\
& \multicolumn{2}{l|}{tweet $\rightarrow$ $V_{sadness}$} & 0.243 & 0.19 & 0.9471 & 0.5554 & 0.8649 \\
& \multicolumn{2}{l|}{combined} & 0.2618 & 0.1896 & 0.9454 & 0.5805 & 0.8497 \\
& \multicolumn{2}{l|}{separate} & 0.1985 & 0.194 & 0.9673 & 0.5804 & 0.8968 \\
& \multicolumn{2}{l|}{separate+combined} & 0.1985 & 0.194 & 0.9673 & 0.5804 & 0.8968 \\
\hline
Spanish & \multicolumn{2}{l|}{tweet $\rightarrow$ $V_{joy} \cup V_{sadness}$} & 0.2288 & 0.1679 & 0.97 & 0.5356 & 0.822 \\ 
& \multicolumn{2}{l|}{tweet $\rightarrow$ $V_{joy}$} & 0.3764 & 0.1596 & 0.9218 & 0.5061 & 0.8442 \\
& \multicolumn{2}{l|}{tweet $\rightarrow$ $V_{sadness}$} & 0.3259 & 0.1623 & 0.9376 & 0.5051 & 0.8657 \\
& \multicolumn{2}{l|}{combined} & 0.3096 & 0.1634 & 0.944 & 0.5228 & 0.8363 \\
& \multicolumn{2}{l|}{separate} & 0.3685 & 0.1601 & 0.9252 & 0.4986 & 0.8611 \\
& \multicolumn{2}{l|}{separate+combined} & 0.3331 & 0.1627 & 0.9397 & 0.5102 & 0.875 \\
\hline
\end{tabular}
\end{center}
}
\caption{Task 1 valence test set results with separate learning. Combined learning has twice the amount of training data and test data is also doubled accordingly.}
\label{Task1_RTMseparated}
\end{table*}

The top $5$ features selected for Task 10 are listed in \Cref{top5features_Task10}.
$1$\&$2$gram w$F_1$ is $F_1$ score over $1$-gram and $2$-gram features with recall computed according to the sum of the likelihood of 
observing them among $1$-grams or $2$-grams correspondingly (wrec) and precision computed according to all corresponding counts in $n$-grams. wGM is weighted geometric mean of the arguments of $F_1$. The features enable linking $w_1$ and $a$ and $w_2$ and $a$ and $8$ of the top $10$ features use $n$-gram features, which makes sense for linking words and we observe this even semantically for Task 10.
The top $5$ features selected for Task 1 are listed in \Cref{top5features_Task1}. bpw is bits per word. WER is word error rate.

\begin{table*}[t]
{
\begin{center}
\begin{tabular}{@{\hspace{0.0cm}}lll|lllll@{\hspace{0.0cm}}}
\multicolumn{3}{c|}{Task 1} & $r$ & MAE & RAE & MAER & MRAER \\
\hline
Task & Model & & \multicolumn{5}{c}{$a (x_1 - x_2) + b$} \\
\hline
English & \multicolumn{2}{l|}{combined} &  \\
& \multicolumn{2}{l|}{separate} &  \\
\hline
Arabic & \multicolumn{2}{l|}{combined} & 0.044 & 0.2 & 0.997 & 0.6069 & 0.876 \\
& \multicolumn{2}{l|}{separate} & -0.005 & 0.201 & 1.002 & 0.6096 & 0.884 \\
\hline
Spanish & \multicolumn{2}{l|}{combined} & 0.056 & 0.1735 & 1.003 & 0.5539 & 0.841 \\
& \multicolumn{2}{l|}{separate} & -0.05 & 0.175 & 1.011 & 0.5547 & 0.855 \\
\hline
Task & Model & & \multicolumn{5}{c}{$a (x_1 + x_2) / 2 + b$} \\
\hline
English & \multicolumn{2}{l|}{combined} &  \\
& \multicolumn{2}{l|}{separate} &  \\
\hline
Arabic & \multicolumn{2}{l|}{combined} & 0.238 & 0.1908 & 0.951 & 0.5817 & 0.857 \\
& \multicolumn{2}{l|}{separate} & 0.24 & 0.1907 & 0.951 & 0.5778 & 0.854 \\
\hline
Spanish & \multicolumn{2}{l|}{combined} & 0.34 & 0.1641 & 0.948 & 0.5262 & 0.82 \\
& \multicolumn{2}{l|}{separate} & 0.385 & 0.1603 & 0.926 & 0.5134 & 0.82 \\
\hline
\end{tabular}
\end{center}
}
\caption{Task 1 valence test set results with separate learning and prediction combinations.}
\label{Task1_RTMseparated_distance}
\end{table*}

\section{Experiments After the Challenge}
\label{ExperimentsAftertheChallenge}

We recalculated our results using 200 thousand training instances for Task 1, which also helps comparison with previous results~\cite{Bicici:RTM_SEMEVAL} and an advanced IBM2 alignment model after the challenge.
\Cref{TestResultsNew} lists the new results obtained after the challenge. 
The number of results with MRAER larger than 1 decreased to 5.
The MRAER obtained by RTMs in STS in 2016 is $0.73$~\cite{Bicici:RTM:SEMEVAL2016} and in quality estimation task for English to German in 2017 is $0.8$~\cite{Bicici:RTM:WMT2017}.
Translation similarity distances alone can be useful to identify the discriminative power of 
attributes~\Cref{Task10_translation_similarity_distance}.
The predictions for Task 10 were transformed to binary classes by thresholding with optimized thresholds on the training set. Prediction differences are used from the combined model or the separate model.
\Cref{Task1_RTMseparated} show that valence can be better predicted with MTPP towards the translation of the vocabulary of sadness rather than using the union of contrastive vocabulary sets of joy and sadness. Separate learning also improve the performance for English and Spanish valence prediction.
Translation similarity distances are also used for Task 1 in~\Cref{Task1_RTMseparated} where instead of learning a threshold to optimize for $F_1$ score, we fit a linear model, $a x + b$, to the distance between predictions.

\subsection{Symbolic Grounding of Score Thresholds}

We symbolically ground the score thresholds with respect to the statistics of the training set scores and use the corresponding linear mathematical equation in the form of $a \mu + b \sigma + c$ on the test set statistics to obtain a relative test set threshold. We observe up to $1\%$ improvements in the prediction performance in Task 10 (\Cref{ExperimentsAftertheChallenge}).

\subsection{Symbolic Grounding of Prediction Score Statistics}

We symbolically ground the predictions with respect to the statistics of the training set scores and use the corresponding linear mathematical equation on the test set statistics to obtain relative predictions on the test set. 
We observe up to $\ldots\%$ improvements in the prediction performance (\Cref{ExperimentsAftertheChallenge}).

\begin{table}[t]
\begin{tabular}{l|ll}
Metric & ranking error & $r_S$ \\
\hline
r &  0.0564 & 0.6614 \\
MAE & 0.1167 & 0.2997 \\
RAE & 0.1167 & 0.2997 \\
MAER & 0.1518 & 0.0892 \\
MRAER & 0.0947 & 0.4318 \\
rMAER & 0.0893 & 0.464 \\
rMRAER & 0.0869 & 0.4785 \\
\hline
\end{tabular}
\label{metrics_ranking_comparison}
\caption{Ranking errors and $r_S$ with different metrics.}
\end{table}

\subsection{Ranking Results}

We would like to obtain robust sortings on the training set so that the sortings are reflected on the performance sorting on the test set. We compare the ranking differences that incur when we choose a metric to rank both the training and test set results. \Cref{metrics_ranking_comparison} compares the performance metrics we use using $r_S$, Spearman's correlation, and ranking errors, which compute the squared error of normalized ranking differences (\Cref{rankError}). 

\begin{equation}
rankError = \sum_i (\texttt{rank}(i, \mbox{train}) / n - \texttt{rank}(i, \mbox{test}) / n)^2 \label{rankError}
\end{equation}

$r_P$ is Pearson's correlation and $r_S$ is calculated using the same formula of $r_P$ but using rankings where tied ranks are assigned the mean rank~\cite{Kokoska:1985452}. \Cref{spearmancorrApprox} provides an approximate result when there are ties.
$r_S$ provides higher-level information than $r_P$ since we use the rankings obtained.

\begin{align}
r_P & = & & \frac{\displaystyle\sum_{i=1}^n (\hat{y_i} - \bar{\hat{y}}) (y_i - \bar{y})}{\sqrt{\displaystyle\sum_{i=1}^n (\hat{y_i} - \bar{\hat{y}})^2} \sqrt{\displaystyle\sum_{i=1}^n (y_i - \bar{y})^2}} \nonumber \\
r_S & \simeq & & 1 - \frac{6\sum_{i=1}^n d_i^2}{n(n^2 - 1)} \label{spearmancorrApprox}
\end{align}

Both $r_S$ and $r_P$ are invariant to scaling and we can use \Cref{correlationInvariant}~\cite{Lee_Rodgers_1988} for standardized $\hat{y}$ and $y$ where $xy = \frac{1}{2} (x^2 + y^2 - (x - y)^2)$ equivalence and $\sum_i \hat{y_i} = 1$ and $\sum_i y_i = 1$ are used. Since $-1 \leq r \leq 1$, the variance of the difference is in $[0, 4]$. Similarly, $r_P = \frac{1}{2} s^2_{\hat{\textbf{y}} + \textbf{y}} - 1$, and thus correlation and $s^2_{\hat{\textbf{y}} - \textbf{y}}$ or $s^2_{\hat{\textbf{y}} + \textbf{y}}$ can be used to determine each other~\cite{Lee_Rodgers_1988}. \Cref{spearmancorrApprox} is obtained using \Cref{correlationEquation} and assuming that ranks are distinct.
\begin{align}
r_P & = & & \frac{\displaystyle\sum_{i=1}^n \hat{y_i} y_i}{\sqrt{\displaystyle\sum_{i=1}^n \hat{y_i}^2} \sqrt{\displaystyle\sum_{i=1}^n y_i^2}} \;\; \mbox{(for standardized input)} \nonumber \\
& = & & \displaystyle \frac{1}{n} \sum_{i=1}^n \hat{y_i} y_i \nonumber \\
& = & & \displaystyle \frac{1}{2n} \sum_{i=1}^n (\hat{y_i}^2 + y_i^2 - (\hat{y_i} - y_i)^2) \label{correlationEquation} \\
& = & & \displaystyle 1 - \frac{1}{2n} \sum_{i=1}^n (\hat{y_i} - y_i)^2 \\ 
& = & & \displaystyle 1 - \frac{1}{2} s^2_{\hat{\textbf{y}} - \textbf{y}} \label{correlationInvariant}
\end{align}

\begin{figure*}[t]
\begin{minipage}{12cm}
\begin{align}
\mbox{MAER}(\hat{\textbf{y}}, \textbf{y}) & = & & \frac{\displaystyle\sum_{i=1}^n \frac{|\hat{y_i} - 
y_i|}{\lfloor|y_i|\rfloor_\epsilon}}{n} \\
\mbox{MRAER}(\hat{\textbf{y}}, \textbf{y}) & = & & \frac{\displaystyle\sum_{i=1}^n \frac{|\hat{y_i} - 
y_i|}{\lfloor|\bar{y} - y_i|\rfloor_\epsilon}}{n} \\
\mbox{rMAER}(\hat{\textbf{y}}, \textbf{y}) & = & & \frac{1}{n} \displaystyle\sum_{i=1}^n \frac{|\hat{y_i} - 
y_i|}{\lfloor|y_i|\rfloor_\epsilon} f_{<0, \epsilon} \left(\frac{(\hat{y_i} - \bar{\hat{y}})(y_i - \bar{y})}{\sigma_{\hat{y}} \sigma_{y}\lfloor|y_i|\rfloor_\epsilon^2}\right) \\
f_{<0, \epsilon} (x) & = & & \begin{cases}
   \lfloor x \rfloor_\epsilon   & \text{if } x \geq 0 \\
   \lfloor -2 x \rfloor_\epsilon & \text{if } x < 0
  \end{cases}
\end{align} \\
\label{RelativeEvaluationMetrics}
\end{minipage}
\end{figure*}

MAER and MRAER are capped from below\footnote{We use $\lfloor \;.\; \rfloor_\epsilon$ to cap the argument from below to 
$\epsilon$.} with $\epsilon = \mu_{|\hat{\textbf{y}} - \textbf{y}|} / 2$, which is the 
measurement error and it is estimated as half of the mean absolute error or deviation of the predictions from the 
target mean. 
$\epsilon$ represents half of the score step with which a decision about a change in measurement's value can be made. 
$\epsilon$ is similar to half of the standard deviation, $\sigma$, of the data but over absolute differences. 
For discrete target scores, $\epsilon = \frac{\textit{step size}}{2}$.

We use standardized relative correlation measure~\cite{Bicici:RTM:SEMEVAL2016}. rMAER combines correlation with MAER using a reciprocal function of negative values, $f$. rMRAER is defined similar to rMAER.

\section{Conclusion}


Referential translation machines obtain automatic prediction of semantic similarity using MTPP. 
We presented encouraging results with stacked RTM models for predicting the 
intensity of the structure and content in text with our novel MTPP modeling for translation to WordNet emotion lists and the discriminative power of attributes using stacked RTM models.
Our results also enable comparisons of prediction results of RTMs in different natural language processing tasks~\cite{RTMComparison}.

%

%

%

\bibliography{ebiciciPubs,Bibtex}

\pagebreak

\appendix

\section{English Lexicon used from WordNet Affect Emotion Lists}
\label{wordnetlexicon}

English lexicon used for identifying discriminative attributes (Task 10) as a subset of the WordNet affect emotion lists~\cite{Strapparava2004} available at \url{http://web.eecs.umich.edu/~mihalcea/affectivetext/}.
~\footnote{The original lexicon was made available freely for research purposes. The lexicon in this section is for the publication and demonstration.}

{\small
\hspace*{-0.5cm}
\begin{tabular}{@{\hspace{0.0cm}}c@{\hspace{0.0cm}}}
\textit{anger} \\
\parbox{8.75cm}{torment pissed frustrated crucify pout grudge umbrageous incensed spitefulness execration exacerbate choleric infuriating displease teasing aggression misology maliciously riled ire baffled spiteful pestered enragement rag misanthropy misogynism harried grudging annoyance abominate displeased furious irritating wrothful nettlesome detest vexation hatred indignantly angry offense vindictiveness despising revengefully disdain belligerent murderousness stung misanthropical lividly madness irritation belligerently misogynic begrudge abhor temper indignation infuriated anger discouraged grievance contemn bitterness vexatious rancour resentment hate exasperating malign covetous enraged envious enviously jealous irascibility loathing displeasingly gravel odium grasping infuriate sulkiness outrage annoyed mad covetously vindictively belligerence prehensile enviousness scene harassment murderously enfuriate wrathful hateful displeasure loathe scorn maliciousness execrate venom enviably pique roiled heartburning malicious tantrum irritated envy frustration nettle harassed spleen brood enviable offend animosity despiteful maddening sore balked aggressive enmity vexed vex jealously resentfully angered galling pestering vindictive hackles umbrage bothersome score gall misanthropic vexing persecute aggressiveness indignant amuck oppress rancor jealousy annoying resentful covetousness frustrating outraged tantalize devil greedy hostilely wrathfully covet exasperation frustrate hostility misogyny incense rage choler malevolently malignity malevolence vengefulness huffiness malevolent furiously chafe pesky begrudging huffy angrily harass wroth despise fury rile avaricious malefic bother abomination spite nettled antagonism aggravated hatefully aggravate hostile belligerency annoy maddened provoked exasperate malice plaguey vengefully displeasing lividity stew pestiferous evil wrath irritate infuriation grizzle aggravation abhorrence misoneism fit} \\
\textit{fear} \\
\parbox{8.75cm}{atrocious anxiously diffidently coldheartedness cruelty frightening hesitantly chill dreadfully scarily panic foreboding hideous panicky hesitancy timidness hesitance cower intimidation ruthlessness dreaded alarm suspense intimidate chilling dreadful horrify diffidence alert fearful presentiment creep apprehension hysterical uneasily anxious bashfully horribly timid apprehensive ugly hardheartedness cringe unkind browbeaten horrifying fawn shadow unassertiveness horrifyingly dread awfully intimidated frightful afraid apprehensiveness outrageous premonition panicked shy scary frighten hysterically heartlessly pitilessness shyly frighteningly fearfulness diffident monstrously dismay timorous fear horrified timorousness unassertive fright cruelly scared consternation affright apprehensively dire timidity heartlessness heartless timidly scarey hysteria alarmed horrific cowed presage shyness mercilessness awful trepidation hideously unsure scare bullied frightened horridly fearsome crawl creeps direful terrified fearfully suspensive shuddery} \\
\end{tabular}
}

{\small
\hspace*{0.35cm}
\begin{tabular}{@{\hspace{0.0cm}}cc@{\hspace{0.0cm}}}
\textit{sadness} \\
\parbox{8.75cm}{disconsolate sorrow guilt dolorous demoralization weight dispirit remorse oppression heartbreaking sorrowfully drear grievously ruefulness sorry oppressed despondently despondency wretched woefulness dreary desolation bereft downcast attrition forlornly sadness plaintive pitiable woebegone penitently pitying desolate heartsickness demoralize dispiritedness dysphoria depressive loneliness persecuted remorsefully sadly dark joylessly regretful despairingly tyrannical suffering sadden ruthfulness distressed melancholic sad bored discouraged disheartening regret penitentially plaintively heartache forlorn tearfulness lachrymose ruefully cheerlessness forlornness gloomily oppressive contrite shamefaced heartrending drab plaintiveness melancholy gloomful mournfulness penance brokenheartedness saddening repentant dysphoric depress mourning heartsick repentantly heavyhearted doleful aggrieve contritely hapless rue contrition downheartedness depression tyrannous piteous pitiful uncheerful demoralized laden tearful grievous woeful gloomy penitence godforsaken repent penitent rueful pathetic miserably dolefully bad oppressiveness dismay repentance cheerless dingy heavyheartedness joyless grim grief persecute bereaved grieving grieve disconsolateness lamentably unhappiness guilty oppress helplessness weeping misery oppressively deject hangdog heartbreak demoralising woefully depressed uncheerfulness contriteness dispirited joylessness disheartened poor gloominess depressing remorseful down unhappy low weepiness mournful deplorably blue glum despondent harass gloom downhearted dolefulness mournfully demoralizing woe downtrodden compunction shamed sorrowing dolourous misfortunate dismal sorrowful glooming sorrowfulness cheerlessly dispiriting} \\
\textit{joy} \\
\parbox{8.75cm}{kick identification amicably anticipation cheerfully favourably eager close gratifying suspenseful triumphant captivation beaming protectiveness devoted zest ardour satisfy exhilarating occupy preference exciting happily penchant compatibly jubilancy tenderness near exultingly approved satisfactory merry becharm crush admire joy exult jolly admiration adoration complacent pleased eagerly anticipate fondness regard kindhearted proudly happiness entrance rapport enamoredness affectional keenness closeness praiseworthily hilariously festive affect appreciated exultant elation exhilaration protective prideful gloat impress liking unworried pride friendly soothe move thrill gratify enthusiastically catch gleefully joyous excitement teased titillate gleefulness joyousness uproarious goodwill gaiety great worship worry glad favorably comfortably taste exhilarate perkiness uplift empathy satisfying hearten attachment comfort gratifyingly beguile commendable fulfil affection tender fulfillment titillation sunny jocularity stimulating warmth concern cheer enthralling joyfully comfortableness entranced approving mirthfully content comforting jubilantly predilection jocund console euphoria satisfactorily merrily satisfied partiality strike kindly gladfulness enjoy solace good fascination love zealous sympathetically benevolent fulfilment affectionateness devotedness gladdened riotously buoyancy benevolently enthralled adorably beneficent hilarity amorous warmhearted likable weakness belonging carefree avidness carefreeness affective captivate barrack recreate sympathetic protectively euphoriant urge friendliness jubilant comfortable benefic intimacy admirably enthusiasm respect sympathy esteem bang emotive enchant approval charm favour satisfyingly amative fascinating gloating fancy brotherlike hilarious loyalty lovingly capture bewitching satisfiable beneficed compatibility lovesome identify beneficially zeal cheery revel favourable jubilance fulfill happy lovingness gayly brotherhood jollity congratulate complacence satisfaction lightsomeness romantic intoxicate joyously triumph charitable devotion heart ebullient benevolence exhilarated exuberance approve contented captivated exuberant euphoric exultantly festal rush cheerful endearingly tickle jubilation complacency gladsomeness enthusiastic delighted triumphal joyful affectionate gloatingly charmed eagerness empathic amicability amatory flush kid charge screaming rejoice relish entrancing fondly expansively exuberantly contentment inspire fond like suspensive likeable amicable triumphantly expectancy}
\end{tabular}
}

\end{document}